\documentclass{article}
\usepackage{spconf,amsmath,graphicx,cite,booktabs,multirow}

\title{SYNERGY OF MACHINE AND DEEP LEARNING MODELS FOR MULTI-PAINTER RECOGNITION}

\name{Vassilis Lyberatos,  Paraskevi-Antonia Theofilou, Jason Liartis and Georgios Siolas}
\address{National Technical University of Athens, Athens, Greece}
\usepackage{graphicx}
\usepackage{textcomp}
\usepackage{xcolor}
\usepackage{tabularx,adjustbox,booktabs}
\usepackage{colortbl}
\usepackage{multirow}
\usepackage{caption}
\usepackage{subcaption}
\usepackage{mathtools}
\usepackage{makecell}
\usepackage{caption}
\begin{document}
%
\maketitle
\begin{abstract}
The growing availability of digitized art collections has created the need to manage, analyze and categorize large amounts of data related to abstract concepts, highlighting a demanding problem of computer science and leading to new research perspectives. Advances in artificial intelligence and neural networks provide the right tools for this challenge. The analysis of artworks to extract features useful in certain works is at the heart of the era. In the present work, we approach the problem of painter recognition in a set of digitized paintings, derived from the WikiArt repository, using transfer learning to extract the appropriate features and classical machine learning methods to evaluate the result. Through the testing of various models and their fine tuning we came to the conclusion that RegNet performs better in exporting features, while SVM makes the best classification of images based on the painter with a performance of up to 85\%. Also, we introduced a new large dataset for painting recognition task including 62 artists achieving good results.
\end{abstract}


\section{Introduction}
Art as an integral part of human culture has entered the digital age, following the trend of recent years to digitize our world. Large collections of artworks, mainly from museums and galleries, but also private collections, are digitized, published and made available on the internet with the aim not only of preserving and presenting them as creations of human expression, but also as data for processing and analysis using machine learning methods.

The large volume of this type of data, as well as the advanced tools of artificial intelligence impose the need to automate the work of analysis, extraction of features and categorization of certain components of artworks. . Art historians, curators and art experts, in general, are able to identify the particular characteristics of a painting, its creator, the genre, the art movement and the artistic style to which it belongs. Their ability is based both on their perception and memory, as well as on their artistic experience that allows them to connect the various artistic features with each other and to draw relevant conclusions. Therefore, the aim is to imitate the human ability, experience and knowledge to automate the above tasks.

We are interested in the problem of automatic painter recognition. Approaching this problem, we consider that the artworks of a painter present common features related to the style and preferences of each artist. However, it has been observed that an artist does not follow from the beginning to the end of his course the same art movement, the same genre, the same theme, the same style or color palette, as well as may have been influenced by other creators, for example by their teachers, in which case their paintings have common features. Therefore, the present work is called to deal with the complexity of recognizing a painter from a painting by finding appropriate methods in machine learning. 

Recognizing the creator, the genre, the art movement and the style of a painting are problems of increasing complexity, the solution of which is a research challenge in the field of machine learning. Deep learning techniques are mainly employed, but also classical machine learning techniques are used to deal with these complex problems. On the one hand, classical methods of image processing are used for feature extraction and combined with typical machine learning classifiers. On the other hand, feature extraction and classification is done by an end-to-end training of convolutional neural networks and a use of transfer learning. We combined these two approaches in order to yield better results to the painter recognition task. Having image data from the WikiArt repository and knowing the metadata that accompanies the painter in each painting, we fine-tuned pre-trained deep neural networks to extract the appropriate features from images and then used  machine learning classifiers on those features to get our predictions.

The advantage of our method lies in its application to a new large-scale dataset of 62 artists. There are no other studies, as we know, that have approached the specific problem to such an extent, achieving such good results without the use of end-to-end deep neural networks. Thus, we managed to approach the task of painter recognition using a different from the state-of-the-art methodologies that gives satisfactory results for usual datasets of about 20 artists and excellent for the problem of increased complexity of 62 artists.

\section{Background}\label{sec:backround}
Automation in the processing, analysis and categorization of artworks has been a major challenge in recent decades and machine learning has made a significant contribution to progress in this field. There are many studies regarding the distinction of the type and style of a painting, as well as the identification of its artist. 

Early studies addressed these problems with traditional machine learning methods based on low-level features and a relatively small number of artworks \cite{keren2002painter, gunsel2005content, lombardi2005classification, jiang2006effective, zujovic2009classifying, arora2012towards, falomir2018categorizing}. These studies use features that capture shape, texture, edge and color properties and are extracted with the use of classical computer vision methods, like SIFT, GIST, HOG, GLCM, and HSV color histograms. Then, they train classical classifiers of machine learning, like SVM, K-NN, Random Forest, MLP, to make the predictions.

Following the studies, CNN was introduced as extractors of features. The first large-scale such study \cite{karayev} has shown that features derived from the layers of a CNN pre-trained in non-artistic images achieve high performance in the tasks of art classification. Then, many studies \cite{bar, peng, david, saleh, cetinic16, Agarwal2015, crowley, GatysEB15a} have confirmed the effectiveness of the methods based on the extraction of features using CNN, as well as the combination of them with the handmade features of the images.

Regarding the problem of painter recognition from a painting, in which we are interested in this work, in 2013, Cetinic et al. \cite{cetinic2013automated} studied the style of individual artists by extracting specific features, like color, light, texture, and then several classifiers, such as MLP, SVM, Naïve Bayes, Random Forest and Adaboost, were applied. 
In 2015, Saleh et al. \cite{saleh} investigated the applicability of metric learning approaches and performance of different visual features (feature fussion) coupled with SVM for learning similarity between artistic items. 
In 2016, Tan et al. \cite{tan2016ceci} used CNN as feature extractor and SVM for classification or CNN as an end-to-end fine-tuned architecture. In 2017, Viswanathan \cite{viswanathan2017artist} trained from scratch a ResNet18 to resolve the problem of painter recognition. In 2018, Cetinic et al. \cite{cetinic2018fine}, also, focused on fine-tuned networks based on VGG, Resnet, Googlenet and CaffeNet models. In 2019, Kelek et al. \cite{kelek2019painter} used pre-trained architectures such as GoogleNet, Inceptionv3, ResNet50, ResNet101 and DenseNet. Zhong et al. \cite{zhong2020fine} proposed a dual path classification scheme, including RBG and brush stroke information channels, based on architecture of ResNet131.
In 2020, Choundhury \cite{choudhury2020automated} compared the results of feature extraction methods to train Random Forest and SVM classifiers and deep convolutional networks with transfer learning, such as basic CNN, ResNet18 and ResNet50. 
In 2021, Cömert et al. \cite{comert2021painter} focused on the fine-tuning of MobileNet v2, ResNet, Inception v2 and NasNet to identify the painter of an artwork. The same year, Zhao et al. \cite{zhao2021compare} compared the architectures ResNet, RegNet and EfficientNet for painter identification. Finally, Nevo et al. \cite{nevo2022deepartist} proposed a novel dual-stream architecture, based on pretrained EfficientNet model, for capturing in parallel both global elements and local structures in painting's images. It is important to mention that most of the related work has used datasets derived form WikiArt.

\section{Dataset and Features}\label{sec:datasets}

Our data were retrieved from Kaggle\footnote{https://www.kaggle.com/c/painter-by-numbers/overview}. Most of these digitized images were taken from the WikiArt repository. In total, 103.250 paintings by 2.319 artists are available in this dataset. 
It is noted that for our convenience in managing our data and in the application of the various techniques and algorithms that we have chosen, we built two different datasets. We did that in order to test our algorithms in different scales and difficulty of the problem. The first one corresponds to 20 distinct painters with 9,986 paintings in total (\textit{Medium Dataset}) and the second one to 62 with 26,263 (\textit{Large Dataset}). The selection of data was based on the minimum number of paintings for each artist in order to have adequate number of artworks per artist. For this reason, we selected for the study those painters who have at least 270 artworks in the original dataset. In addition, in order to do equal comparison with state-of-the-art works, we applied our experiments to the dataset used by the state-of-the art (SOTA) methods, which contains 19,050 paintings for 23 painters \footnote{https://github.com/cs-chan/ArtGAN} and is derived from WikiArt repository. From this point we will refer to this dataset as \textit{SOTA Dataset}.




Furthermore, for the best management of this two dimensional data, resizing of the images (3x256x256) was applied, so that the dimensions of the input to our algorithms are specific. For the better generalization of our models we performed data-augmentation, we used random cuts and reflections of the images. We also applied normalization to the data based on the mean value and the covariance of the features, which helped to accelerate the training of our neural networks.

\section{Methodology}

Regarding the feature extraction from images, none of the classic techniques were applied, but it was preferred to use SOTA pre-trained convolutional neural networks that are distinguished for their ability to extract high-level features from images. This is also a challenge of our study, as it is interesting to investigate to what extent convolutional networks can offer to classification networks those appropriate features that will help to solve the problem of painter recognition as effectively as possible.

To construct a deep-neural-network-based feature extractor we conducted many experiments with different SOTA architectures and datasets. We tried different sizes of ResNet\cite{he2016deep} and RegNet\cite{radosavovic2020designing} architectures in different datasets. We started by experimenting on the \textit{Medium dataset} in order to select an architecture. The RegNet architecture outperformed the ResNet, so we continued experimenting with it on the \textit{Large dataset}. 

Further experiments focused on determining the best hyper-parameters for the RegNet architecture. We experimented with different ways of training. We tried different values of \textit{model depth} and \textit{learning rate} in order to a model of the right capacity. Additionally we tried adding a \textit{dropout} layer. Also, we used new techniques such as \textit{label smoothing}\cite{NEURIPS2019_f1748d6b}, \textit{warm-up layers} and \textit{frozen layers} in order to avoid over-fitting and catastrophic forgetting \cite{FRENCH1999128}. 

After choosing the best architecture for each sub-task we applied various machine learning algorithms. We applied heuristic algorithms for finding the best combination of hyper-parameters for each model and ended up selecting the best one. We followed the same procedure for the \textit{Large Dataset} and we tried one extra SOTA classifier, \textit{XGBoost}. Also in order to compare with SOTA model's we applied our best model on the \textit{SOTA Dataset}. Our aim was to use classifiers that don't follow the stantard MLP model with fully connected layers. 

A crucial factor for achieving good results was the hyper-parameter tuning. Specifically for our best classifier \textit{SVM} we tuned the hyper-parameters: \textit{C}, \textit{$\gamma$} and \textit{kernel-type}. We performed gradually grid-search in order to find the best combination for our model, using the sklearn python package.


\section{Experiments}

Our models managed to be comparable to SOTA models in the painter recognition task, especially for the works due to 2021, where started approaching the problem with different manner using dual-stream architectures for feature extraction. In Table~\ref{tab:sota} the performance of our best models is given compared to previous works. For the implementation of our experiments we provide a GitHub repository~\footnote{https://github.com/jliartis/art-recognition}. Below our experiments are described analytically.

\subsection{Recognition of 20 artists}\label{20_artists}

The sub-task of recognizing paintings from 20-artists was based in the \textit{Medium Dataset}. We conducted a number of experiments in order to choose the best architecture for feature extraction. We split the dataset in three parts: 20\% validation set, 20\% test set and 60\% train set. We experimented with different sizes of the layers of ResNet (34, 101, 152) and of the number of parameters of RegNet (400mf, 800mf). We had approximately 20 experiments and ended up choosing RegNet\_Y\_800MF and ResNet-152. After choosing the two best architecture we did fine tuing to choose one of them. RegNet\_y\_800MF with \textit{learning rate} 1e-5, trained in 20 \textit{epochs} with 3 \textit{frozen layers} had the best results achieving 0.84 accuracy in the validation set, so we chose it for our feature extractor.

After using the extractor to retrieve the desired features, we applied 6 different ML classifiers: \textit{Logistic Regression} , \textit{Naive Bayes}, \textit{K-Nearest Neighbors} (K-NN), \textit{Decision Trees}, \textit{AdaBoost} and \textit{Support Vector Machines} (SVM). We applied cross-validation with 5 \textit{folds}. In order to compare them fairly, we applied hyper-parameter optimization to all of them, using the gridsearch algorithm. After conducting all the experiments \textit{SVM} prevailed all the others reaching 85\% accuracy, while \textit{Decision Trees} having the worst results. An overview of the results is shown in Table~\ref{tab:medium_classifiers}.

\begin{figure}[!b]
    \centering
    \tiny
    \includegraphics[scale=0.35]{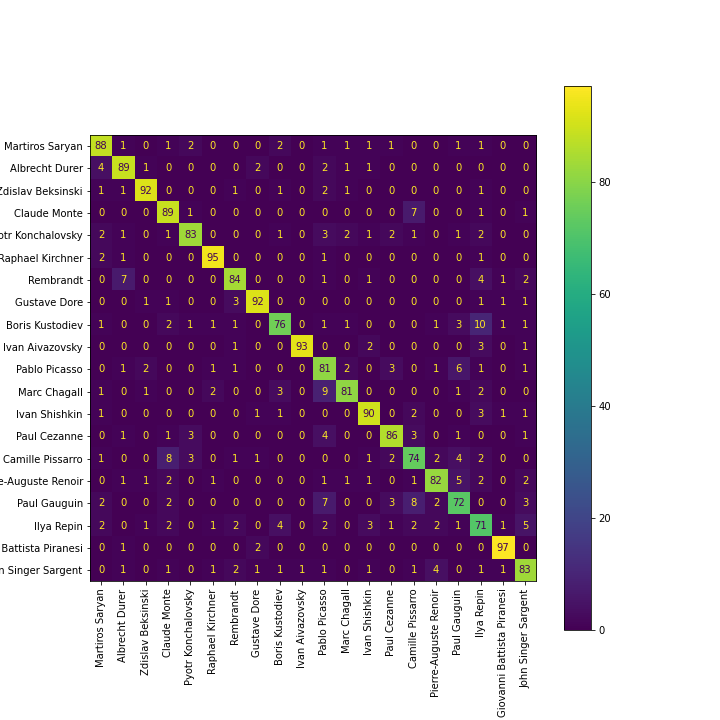}
    \caption{Confusion matrix for the best SVM trained in a subset of the \textit{Medium Dataset}}
    \label{fig:confusion}
\end{figure}

From Fig.~\ref{fig:confusion}, we can conclude that in general the performance of our method was good for all classes. We had the worst results for the painter Ilya Repin and the best for the painter Battista Piranesi. It is also noted that there was a confusion between the painters Boris Kustodiev and Ilya Repin. This confusion may be due to the fact that both artists belong to the same artistic movement, Realism. In addition, both are from Russia and studied at the same time at the Imperial Academy of Arts of Russia.


\begin{table}[!b]
    \caption{Comparing our ML classifiers}
    \centering
    \small
    \begin{tabular}{m{0.3\columnwidth}>{\centering\arraybackslash}m{0.25\columnwidth}>{\centering\arraybackslash}m{0.21\columnwidth}}
    Classifier & \textit{Medium Dataset} & \textit{Large Dataset}  \\\midrule
    Plain RegNet & - & 79\%\\
    Decision Trees & 63\% &  -\\
    Random Forest & - & 69\% \\
    AdaBoost & 79\% & - \\
    Naive Bayes & 82\% & -\\
    Logistic Regression & 84\% & -\\
    K-NN & 84\% & 72\%\\
    XGBoost & - & 75\%\\
    \textbf{SVM} & \textbf{85\%} & \textbf{84\%}\\
    \end{tabular}
    \label{tab:medium_classifiers}
\end{table}



\subsection{Recognition of 62 artists}

The second sub-task with the 62-artists is based on the \textit{Large Dataset}. We split the dataset in three parts: 20\% validation set, 20\% test set and 60 \% train set. Since the best model in our previous experiments (see Section~\ref{20_artists}) was RegNet architecture we chose it a priori. In order to chose the best hyper-parameters for RegNet we had 30 different experiments. We tried various sizes of the model's parameters (400mf, 800mf, 1.6gf) and ended up choosing RegNet\_Y\_1\_6GF. The hyper-parameters of the best model that achieved 0.8 accuracy in the validation set are: 1000 \textit{epochs}, 128 \textit{batch size}, 4 \textit{warm up layers}, 2 \textit{freeze layers}, 0.015 \textit{label smoothing} and 1.6 \textit{learning rate}.

Since we chose the best feature extractor for this task, we tried four machine learning classifiers for the task: \textit{Random Forest}, \textit{K-Nearest Neighbors}, \textit{Support Vector Machines} and \textit{XGBoost}. We used cross-validation for the training and tuning of each classifier with 5 \textit{folds}. For the hyper-parameter tuning of the classifiers we used Optuna~\cite{akiba2019optuna}, a python library. The best classifier was proven to be \textit{SVM} achieving 84\% accuracy and the worst was \textit{Random Forest} achieving 69\%. The results are displayed in Table~\ref{tab:medium_classifiers}. It is important to point that plain RegNet achieved 79\%, which means that SVM boosted our performance with 5 percentage points.  

\subsection{Comparison with state-of-the-art models}

In order to compare our methodology with SOTA \cite{saleh}, \cite{cetinic2018fine}, \cite{zhao2021compare}, \cite{nevo2022deepartist}, we applied our experiments to the same dataset, the performance in which can be a mark for the efficiency of our approaching system. We used our RegNet model that was fine-tuned on the \textit{Large Dataset} for extracting representative features of the dataset and then performed gridsearch for SVM models. 

We achieved results close to SOTA model's performance as shown in Table~\ref{tab:sota}. Our approach deviates from the so far proposed end-to-end deep convolutional network architectures to solve the task of painter recognition and proves that traditional classifiers have still the potential to have comparable results to pure deep-learning models.

\begin{figure}[!t]
    \centering
    \tiny
    \includegraphics[scale= 0.5]{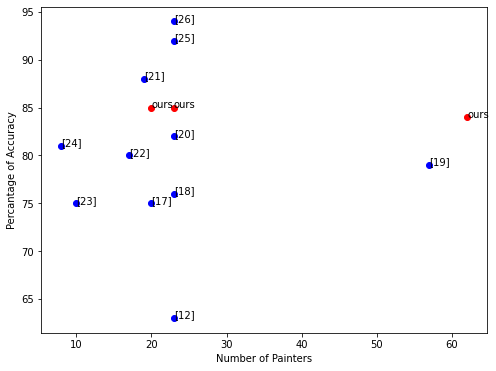}    \caption{Scatter plot for comparing the change in accuracy with the number of artists (blue points represent previous works and red ones ours -  all points are annotated with the corresponding references) }
    \label{fig:scatter}
\end{figure}

We consider that that the complexity of the problem lies in the increase in the number of artists to be recognized. Regarding Fig.~\ref{fig:scatter} we observe that our methodology achieves appreciable performance for the ordinary datasets and much better for the extended dataset of 62 artists. The references of Fig.~\ref{fig:scatter} are briefly descibed in Section~\ref{sec:backround}.

\begin{table}[!b]
    \centering
    \small
    \caption{Comparing our results with previous state-of-the-art models on \textit{SOTA Dataset}}
    \resizebox{\columnwidth}{!}{
    \begin{tabular}{lclccc}\\\toprule
    Work & Year & Methodology  & Accuracy  \\\midrule
    Saleh et al. \cite{saleh} & 2015 & Feature fussion 
    \&\ SVM & 63\%\\
    Cetinic et al. \cite{cetinic2018fine} & 2018 & CaffeNet   & 82\%\\
    Zhao et al.\cite{zhao2021compare} & 2021 & EfficientNet &  92\%\\
    Nevo et al.\cite{nevo2022deepartist} & 2022 & Dual-stream 
    (EfficientNet)  & 94\%\\
    \textbf{Our approach} & \textbf{2023} & \textbf{RegNetY-1.6MF 
    \&\ SVM} & \textbf{85\%}\\  
    \bottomrule
    \end{tabular}
    }
    \label{tab:sota}
\end{table}

\section{Conclusions and future work}

In this work, we had a large number of experiments trying a variety of SOTA deep learning and machine learning architectures achieving very good results in the painters recognition task. A new methodology was applied, as a synergy of old and new techniques, in order to reach high accuracy in a large-scale and complex multi-class classification problem, like this of 62 artists. It is proven that traditional machine learning classifiers can have close to SOTA results, assembling more versatile model-structures. The use of a pre-trained deep neural network for feature extraction and an SVM for classification instead of the use of an end-to-end deep neural network has advantages in the performance and can lead to a different comparable approach with SOTA ones.

Future work will be the extension of our experiments in other art recognition tasks in order to success the generalization of our model. One other interesting  future direction, in order to tackle the problem of generalization, is to try multi-task learning. Finally, it will be useful to include explainability to make the artist identification process more understandable and detect possible bias of model's parameters. 


\clearpage

\bibliographystyle{IEEEtran}
\bibliography{mybibfile}

\end{document}